\documentclass[letterpaper]{article} 
\usepackage{aaai25}  
\usepackage{times}  
\usepackage{helvet}  
\usepackage{courier}  
\usepackage[hyphens]{url}  
\usepackage{graphicx} 
\usepackage{natbib}  
\usepackage{caption} 
\usepackage{algorithm}
\usepackage{algorithmic}
\usepackage{booktabs}
\usepackage{multirow}
\usepackage{amssymb}
\usepackage{amsmath}
\usepackage{pifont}
\usepackage{subfiles}
\usepackage{newfloat}
\usepackage{listings}
\DeclareCaptionStyle{ruled}{labelfont=normalfont,labelsep=colon,strut=off} 
\floatstyle{ruled}
\newfloat{listing}{tb}{lst}{}
\floatname{listing}{Listing}
\title{World Knowledge-Enhanced Reasoning Using Instruction-guided Interactor\\ in Autonomous Driving}
\author{
    Mingliang Zhai\textsuperscript{\rm 1,2,3}, 
    Cheng Li\textsuperscript{\rm 1,3}, 
    Zengyuan Guo\textsuperscript{\rm 3}, 
    Ningrui Yang\textsuperscript{\rm 1,3}, 
    Xiameng Qin\textsuperscript{\rm 3},  \\
    Sanyuan Zhao\textsuperscript{\rm 1}, 
    Junyu Han\textsuperscript{\rm 3}, 
    Ji Tao\textsuperscript{\rm 3}, 
    Yuwei Wu\textsuperscript{\rm 2,1}\thanks{Corresponding author: Yuwei Wu},
    Yunde Jia\textsuperscript{\rm 2,1}
}
\affiliations{
    \textsuperscript{\rm 1}Beijing Institute of Technology \\
    \textsuperscript{\rm 2}Shenzhen MSU-BIT University\\
    \textsuperscript{\rm 3}Chongqing Changan Automobile Co., Ltd.\\
    \{zhaimingliang, licheng, ningrui, zhaosanyuan, wuyuwei, jiayunde\}@bit.edu.cn \\
    \{guozy, qinxm, hanjunyu, taoji\}@changan.com.cn
}

\begin{document}

\maketitle





\begin{abstract}
The Multi-modal Large Language Models (MLLMs) with extensive world knowledge have revitalized autonomous driving, particularly in reasoning tasks within perceivable regions. 
However, when faced with perception-limited areas (dynamic or static occlusion regions), MLLMs struggle to effectively integrate perception ability with world knowledge for reasoning. 
These perception-limited regions can conceal crucial safety information, especially for vulnerable road users.
In this paper, we propose a framework, which aims to improve autonomous driving performance under perception-limited conditions by enhancing the integration of perception capabilities and world knowledge.
Specifically, we propose a plug-and-play instruction-guided interaction module that bridges modality gaps and significantly reduces the input sequence length, allowing it to adapt effectively to multi-view video inputs.
Furthermore, to better integrate world knowledge with driving-related tasks, we have collected and refined a large-scale multi-modal dataset that includes 2 million natural language QA pairs, 1.7 million grounding task data. 
To evaluate the model’s utilization of world knowledge, we introduce an object-level risk assessment dataset comprising 200K QA pairs, where the questions necessitate multi-step reasoning leveraging world knowledge for resolution. 
Extensive experiments validate the effectiveness of our proposed method.

\end{abstract}

\section{Introduction}

\begin{figure}[t]
    \centering
    \includegraphics[width=\linewidth]{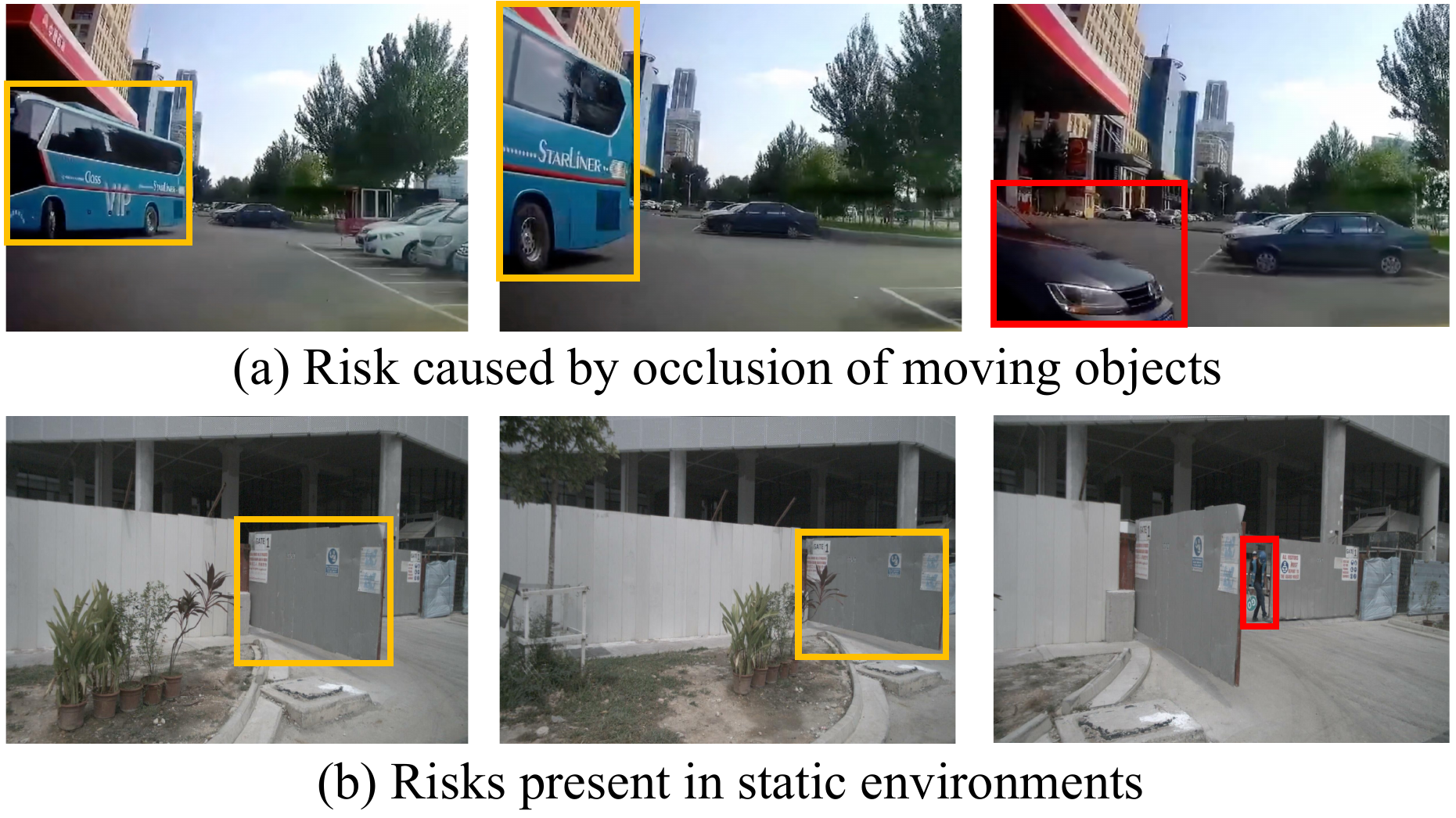}
    \caption{
        \textbf{Examples of dynamic and static environments risks.} 
        (a) The bus in motion severely obstructs the line of sight, resulting in the black sedan being hidden, which significantly increases the risk of a traffic accident in an unprotected scenario. 
        (b) Buildings in static scenes can also become occluding objects. For example, in a construction site scene, the construction gate blocks the workers behind the gate.
    }
    \label{fig:examples}
\end{figure}

The Multi-modal Large Models (MLLMs) alleviates the limitations of expert knowledge and training data diversity in traditional autonomous driving systems. 
Recent research~\cite{wen2023dilu, ma2023dolphins, DriveVLM, chen2024drivingwithllms, sima2023drivelm, cui2023drivellm, wang2024omnidrive, ding2024holistic, bai20243d, tian2024tokenize} have made significant progress in understanding and reasoning about perceivable regions. However, there remain deficiencies in handling perception-limited regions, \textit{e.g.}, occluded areas caused by dynamic or static obstacles such as bus and buildings.
As shown in Figure~\ref{fig:examples}, autonomous driving systems typically plan and control only within the perceived areas, while hidden potential risks are critical factors leading to severe accidents. These occluded areas may conceal information crucial to road safety, especially for undetected vulnerable road users, such as pedestrians and cyclists, who are particularly susceptible to the effects of these occlusions. 
We consider that a promising solution is to utilize instruction-guided extraction of highly aggregated visual embeddings to fully leverage the world knowledge encoded in multi-modal large language models for inference.

Currently, methods utilizing MLLMs for driving tasks are primarily categorized into the following three types:
1. Fine-tuning MLLMs ~\cite{wang2024omnidrive, ding2024holistic, wen2023dilu, sima2023drivelm, cui2023drivellm, fu2024drive} directly for tasks such as prediction and planning.
2. The dual-branch system ~\cite{DriveVLM, ding2023hilm, mei2024continuously} for separating and managing tasks based on real-time requirements, addressing time constraints with fast and slow branches. 
3. The training-free method ~\cite{dewangan2023talk2bev, wang2024drivecot, ma2023dolphins} based on the chain of thought. 
These three types of methods have shown promising results, but there are two main issues.
Firstly, MLLMs are not well-suited for multi-view video inputs, which limits the model’s ability to fully leverage perception ability and integrate world knowledge into subsequent reasoning processes.
Secondly, due to the constraints on the input sequence length of MLLMs, aligning inputs with widely used autonomous driving systems is challenging.


In this paper, we propose a multi-view multi-modal unified architecture, which aim to integrate perception ability and world knowledge. 
The core of the architecture is the instruction-guided interaction module to adapt multi-view video inputs and enhancing the correlation between visual features and natural language instructions, facilitating pre-fusion of features across views and modalities. 
We select the $\operatorname{top-k}$ most similar visual features as visual queries, integrating these queries with original visual features using a cross-attention mechanism to generate enhanced and highly aggregated visual representations. 
This pre-fusion strategy not only aids subsequent decoders in more efficient inference but also significantly reduces the length of input sequences, thereby adapt to the inputs of autonomous driving systems. 

To align between multi-view video feature and language embedding space, we collected and refined a large-scale visual-textual dataset aimed at supporting highly complex scene understanding and response capabilities. 
This dataset comprises over 1.7 million annotated location entries and 2 million dialogue records, covering a diverse range of real-world scenarios. 
Furthermore, to address specific corner cases, we employ GPT-4o (for multi-modal information extraction) and GPT-4o-mini (for pure text reasoning path generation), selecting challenging scenarios from NuScenes such as occlusions, traffic violations, and potential collision risks. 
For these scenarios, we conduct thorough object-level risk assessments.
Based on these efforts, we have design a dataset of 200K QA pairs for training a deeper understanding of complex scenes and to evaluate reasoning abilities in perception-limited regions.

In summary, our approach aims to leverage instruction-guided visual embeddings to handle multi-view video data inputs, enhancing the integration of perception ability and world knowledge, and achieving autonomous driving under constrained perception conditions.
Our contribution can be summarized as follows:

\begin{itemize}
    \item We propose a multi-modal large language model architecture tailored for autonomous driving systems, which enhances the perception ability of MLLMs and integrates world knowledge to enable reasoning in perception-limited regions.
    \item We introduce a plug-and-play instruction-guided interaction module that employs a pre-fusion strategy to generate highly aggregated visual features. This module not only facilitates more efficient inference processes in subsequent decoders but also significantly reduces the input sequence length.
    \item We have reorganized the existing datasets for align between multi-view video feature and language embedding space, and propose an object-level risk assessment dataset for evaluating inference performance in perception-limited scenarios.
\end{itemize}

\begin{figure*}[t]
    \centering
    \includegraphics[width=\linewidth]{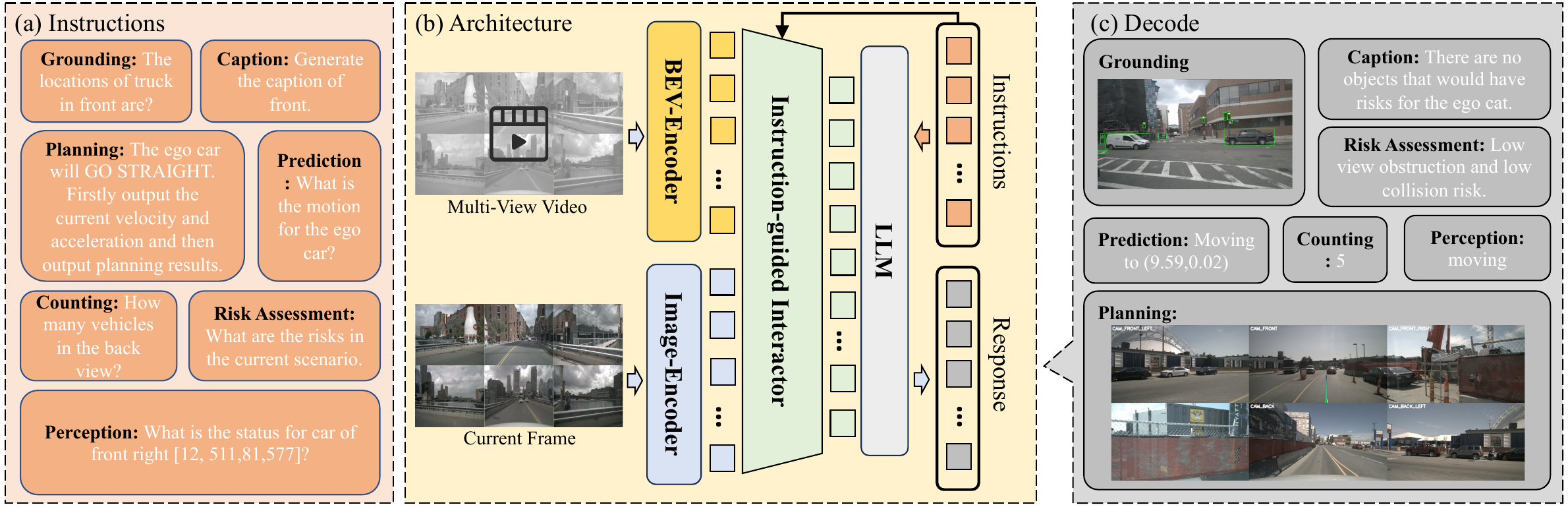}
    \caption{\textbf{Overall of our architecture.} (a) Task-specific instructions. (b) A multi-modal large language model equipped with an interactor, which can select important tokens and perform pre-fusion of these tokens before inputting multi-view and multi-modal information into the LLM. (c) Decoding results and visualization of tokens output by LLM.}
    \label{fig:architecture}
\end{figure*}

\section{Related Works}

\subsection{MLLMs with World Knowledge}
Existing Large Language Models (LLMs) have demonstrated extensive world knowledge~\cite{yukola}, which plays a crucial role in multi-hop reasoning tasks. 
Certain LLMs, such as GPT-4~\cite{achiam2023gpt}, ChatGLM2~\cite{glm2024chatglm}, and LLaMA~\cite{touvron2023llama}, exhibit strong performance on knowledge-driven tasks. 
Recently, MLLMs have introduced world knowledge into the multi-modal domain. 
Some MLLMs, like CLIP~\cite{radford2021learning} and ALIGN~\cite{cohen1997align}, use contrastive learning to create similar embedding spaces for language and vision. 
On one hand, models like LLaVa~\cite{liu2024visual}, PaLM-E~\cite{driess2023palm}, PaLI~\cite{chen2022pali}, RT2~\cite{brohan2023rt}, and InternVL~\cite{chen2024internvl} align images and text tokens using self-attention by interweaving or concatenating tokens of fixed sequence length. 
On the other hand, models such as Flamingo~\cite{alayrac2022flamingo}, Qwen-VL~\cite{bai2023qwen}, and BLIP-2~\cite{li2023blip} employ static queries for cross-attention with visual features to extract a fixed number of visual tokens. 
These approaches effectively map visual features into the linguistic space to leverage world knowledge for reasoning.
However, the utilization of world knowledge is often language-based, and when dealing with multi-perspective video data, visual tokens dominate the input token sequence, thereby diminishing the exploitation of world knowledge. 
We propose a world-knowledge-enhanced MLLM architecture that aggregates visual tokens effectively and maximizes the utilization of world knowledge.


\subsection{MLLMs for Driving Tasks}
For driving tasks, multi-view images or videos are typically required as input.
Approaches for handling multiple image inputs can be categorized into image feature fusion~\cite{awadalla2023openflamingo,laurenccon2024matters,lin2024vila} and image concatenation~\cite{jiang2024mantis,sun2024generative}.
The former approach significantly reduces the resolution of the input images, leading to a loss of image details. The latter approach substantially increases the input sequence length.
Our model adopts a novel approach where relevant features are extracted based on user instructions, and potential details lost are supplemented from the original features.

Previous work typically fine-tunes existing MLLMs with driving  tasks. 
Most existing MLLMs are optimized primarily for visual understanding. 
As a result, autonomous driving MLLMs fine-tuned using these models~\cite{sima2023drivelm, wang2023drivemlm, ma2023dolphins, ding2024holistic, DriveVLM, tian2024tokenize, bai20243d} often lack fundamental 3D understanding and behavioral reasoning capabilities. 
Recent work~\cite{wang2024omnidrive} has integrated detection heads into query transformer. The latent queries used for token extraction also interact with detection queries to guide the tokens to capture 3D perception information. 
However, for perception-limited regions, it is necessary not only to achieve comprehensive perception of the current scene but also to integrate world knowledge for reasoning.

\section{Method}

\subsection{Architecture}
As illustrated in Figure~\ref{fig:architecture}, our overall architecture comprises four key components: 
(1) a shared visual encoder $f_{enc}$, 
(2) a BEV encoder $f_{bev}$,
(3) a instruction-guided interactor $f_{interact}(\cdot) $ that extracts relevant visual tokens based on user requests, and 
(4) a large language model (LLM) $f_{LLM}(\cdot)$ to receive visual and language instruction tokens to generate the response. 

We input multi-view video sequence $\textbf{V}=\{V^i\}_{i=0}^{N_{view}}=[v_1^i, v_2^i, v_3^i, \ldots, v_n^i]$, where $N_{view}$ is the number of views (total 6 views), $n$ is the number of frames.
For clarity in the following, we use $\textbf{L}_{inst}\in \mathbb{R}^{N_{inst}\times D}$ and $\textbf{L}_{resp}\in \mathbb{R}^{N_{resp}\times D}$ to denote the language instruction tokens and response tokens respectively, where $D$ denotes hidden size, $N_{inst}$ and $N_{resp}$ are numbers of tokens for the instruction and response.
We first extract BEV features $\textbf{F}_{bev}\in \mathbb{R}^{N_{bev}\times D}$ by BEV encoder $f_{bev}$, and current frame multi-view image features $\textbf{F}_{mv}=\{F^i_{mv}\}_{i=0}^{N_{view}}\in \mathbb{R}^{N_{mv}\times D}$. 
Notably, after extracting visual features using the encoder $f_{enc}$, we employ an MLP to project the feature dimensions of the visual features to the feature dimension $D$ of the language embeddings.
And then we can formula our architecture as 
\begin{equation}
\label{eq:overall}
    \textbf{L}_{resp} = f_{LLM}\biggl(\textbf{L}_{inst}, f_{interact} \Bigl( \bigl( \textbf{F}_{mv}, \textbf{F}_{bev} \bigr), \textbf{L}_{inst} \Bigr) \biggr).
\end{equation}

\begin{figure}[t]
    \centering
    \includegraphics[width=\linewidth]{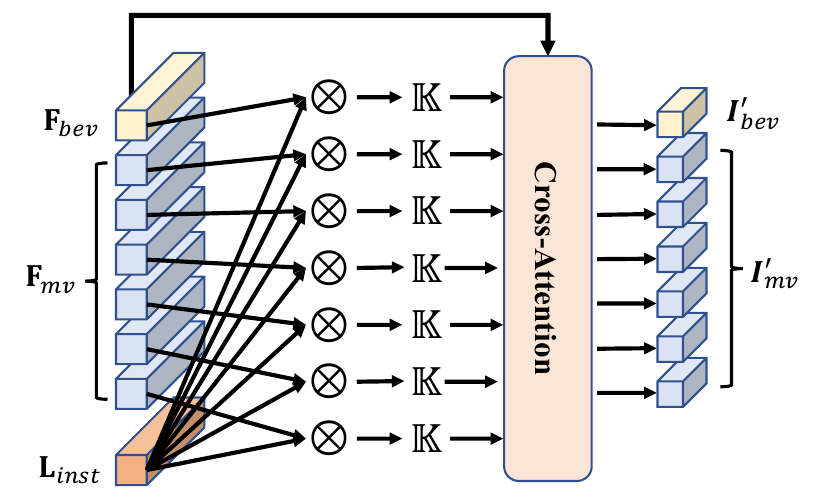}
    \caption{\textbf{Interactor Module.} $\bigotimes $ represents similarity operator. $\mathbb{K}$ represents the $\operatorname{top-k}$ operator.}
    \label{fig:interactor}
\end{figure}

\paragraph{Instruction-guided Interactor.}
Current MLLMs often concatenate information from different modalities directly as input, and then utilizing the global attention mechanism in LLMs to interact with this information. However, the redundant multi-modal tokens can make it challenging for these models to identify useful information relevant to the task. Moreover, as the number of input images or modalities increases, the excessively long input sequences can lead to computational demands that are unacceptably high. This issue is particularly prominent in autonomous driving systems, which require inputs from multiple perspectives and modalities.

To address this issue, we propose the instruction-guided interactor, which can select important tokens and pre-fuse multi-view, multi-modal information before feeding it into LLM. As shown in Figure~\ref{fig:interactor}, the instruction-guided interactor consists of two operations: a selection operation to identify the $k$ tokens most relevant to the language instruction, and an interaction operation to facilitate interaction between the selected tokens and the original features.
The process of the instruction-guided interactor is formulated as 
\begin{equation}
    \begin{aligned}
        F_{mv}^{i'}&=\mathbb{K}\Bigl(\bigl(F_{mv}^i \bigotimes \textbf{F}_{inst}\bigr)\Bigr), \\
        \textbf{F}_{mv}'&=\{F_{mv}^{i'}\}^{N_{view}}_{i=0}, \\
        \textbf{F}'_{bev}&=\mathbb{K}\Bigl(\bigl(\textbf{F}_{bev}\bigotimes \textbf{F}_{inst}\bigr)\Bigr),
    \end{aligned}
\end{equation}
where $\mathbb{K}$ represents the $\operatorname{top-k}$ operator, $\bigotimes$ denotes the computation of similarity between two matrices.
Simply selecting $k$ relevant tokens may result in the loss of some critical information. 
Therefore, inspired by Q-former~\cite{li2023blip}, we enhance the features by computing cross attention between these $k$ tokens and the global features, which can be represented as
\begin{equation}
    \begin{gathered}
    I_{mv} = \operatorname{CrossAttn}(\textbf{F}'_{mv}, \textbf{F}_{mv}, \textbf{F}_{mv}), \\
    I_{bev} = \operatorname{CrossAttn}(\textbf{F}'_{bev}, \textbf{F}_{bev}, \textbf{F}_{bev}),
    \end{gathered}
\end{equation}
where $\operatorname{CrossAttn(\cdot,\cdot,\cdot)}$ is the standard cross-attention operation with the parameters query, key, and value, respectively, $I_{mv}$ and $I_{bev}$ are concatenated and fed into the LLM. 
Notably, the instruction-guided interactor is a plug-and-play module that can be easily extended to more modalities.

\begin{figure}[t]
    \centering
    \includegraphics[width=\linewidth]{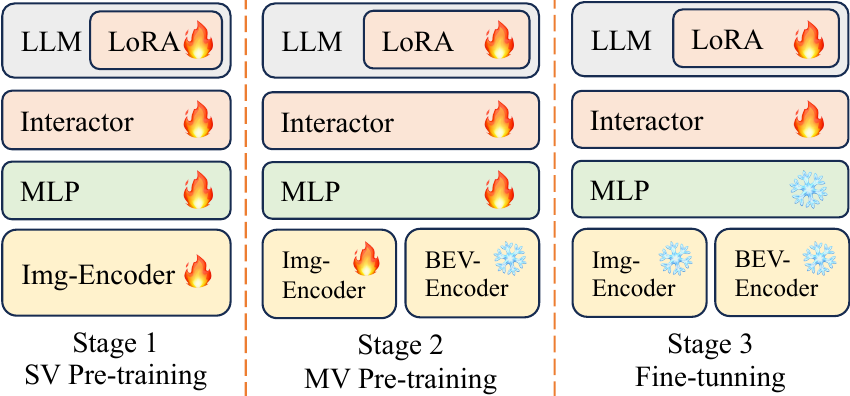}
    \caption{Training pipeline of our method. SV means single-view, and MV denotes multi-view.}
    \label{fig:training_pipeline}
\end{figure}

\subsection{Training Strategy}
Current MLLMs struggle to adapt to multi-view inputs in driving scenarios. To address this issue, we propose a three-phase training strategy. The first phase focuses on aligning the visual and linguistic feature spaces. The second phase is dedicated to constructing relationships between multi-view inputs. The third phase involves instruction fine-tuning to adapt to downstream tasks.
We trained the model following the pipeline shown in Figure~\ref{fig:training_pipeline}.

\paragraph{Stage 1: Single-view Pre-train.}
In this stage, we train our model on single images for captioning and grounding tasks, aiming to establish image-level, region-level, and object-level visual-language alignment. During this process, we unfreeze all parameters except LLM and utilize LoRA to train the LLM.

\paragraph{Stage 2: Multi-view Alignment Pre-train.}
To endow the MLLM with the capability to comprehend multi-view driving scenarios, we extended the dataset from the first stage to incorporate multiple views for model training and incorporated BEV features to provide global semantic information. 
In this phase, the trainable parameters are similar to those the first phase, and the BEV encoder is frozen.

\paragraph{Stage 3: Task-specific Instruction Tuning.}
We have integrated and cleaned multiple open-source datasets. We format all data to Llava's style and use LoRA fine-tune. After this training phase, we obtained a MLLMs capable of engaging in dialogues and exhibiting proficient performance across various driving tasks.

\begin{table}[t]
\centering
\caption{Details of pre-train datasets}
\begin{tabular}{ccc}
\toprule
Task & Pairs \\
\midrule
Grounding-NuScenes & 1700k \\
Caption-NuScenes & 100k \\
\midrule
Total & 1800k \\
\bottomrule
\end{tabular}
\label{tab:ftdata}
\end{table}

\begin{table}[t]
\centering
\caption{Details of fine-tune datasets}
\begin{tabular}{ccc}
\toprule
Dataset & Train & Test \\
\midrule
NuScenes-QA & 376k & 83k \\
NuScenes-MQA & 1204k & 255k \\
OmniDrive-NuScenes & 486k & 90k \\
NuInstruct & 72k & 15k \\
RiskAssessment & 166k & 35k \\
\midrule
Total & 2304k & 478k \\
\bottomrule
\end{tabular}
\label{tab:ftdata}
\end{table}

\section{Dataset Construction}
To achieve multi-modal alignment, we collected and refined a large-scale multi perspective image text pair, including 1.7M grounding data, 200K object-level caption data (objects, risks, weather etc.), 4 open-source datasets and our object-level risk assessment dataset, total 4M samples. Then we format all the data into a unified format. 
Regarding the grounding data, we use a pre-trained Grounding-DINO~\cite{liu2023grounding} model, specifically trained on traffic scenes, to extract all significant objects from single-view images, such as vehicles, pedestrians, traffic signs, and traffic lights.

\paragraph{Object-level Risks Assessment (ORA)}
To evaluate the model performance in perception-limited regions, we propose an object-level risks assessment dataset base on NuScenes~\cite{nuscenes2019}.
We define four types of object-level risks: 1. View obstruction. 2. Collision possibility. 3. Traffic rule violations. 4. Potential risk. 
We classify the QA pairs into six categories: \textbf{Exist} determines whether there is a risk. \textbf{Level} classifies the risk into three levels—low, medium, and high. \textbf{Category} specifies one of the four risk categories mentioned earlier. \textbf{Object} identifies the category of the target causing the risk. \textbf{Reason} describes the cause of the risk. \textbf{Grounding} denotes the location of the target causing the risk.

We use GPT-4o and GPT-4o-mini to construct object-level risk assessment data. 
The construction process is divided into two steps: 
\textbf{Step 1.} We input images along with detailed object information—including category, direction relative to the vehicle, and distance from the vehicle into GPT-4o. We also specify the desired output format to obtain raw data that captures the object-level risks associated with the scene.
\textbf{Step 2.} The raw data generated by GPT-4o is then processed by GPT-4o-mini. This model is used to organize the data and create diversity question-answer pairs that cover different aspects of the object-level risks identified.
The specific prompts and data samples are provided in the appendix.

\begin{table*}[htb]
{
\centering
\caption{Results on ORA dataset.}
\begin{tabular}{c|cccc|cccc|c}
\toprule
\multirow{2}{*}{METHOD} & \multicolumn{4}{c|}{Language Score} & \multicolumn{4}{c|}{Accuracy} & \multirow{2}{*}{mAP $\uparrow$}\\ 
 & BLUE1 $\uparrow$ & BLUE4 $\uparrow$ & CIDEr $\uparrow$ & ROUGE\_L $\uparrow$ & exist $\uparrow$ & level $\uparrow$ & cate $\uparrow$ & object $\uparrow$  & \\
\midrule
Bunny-Llama3        & 62.74 & 39.86 & 244.10 & 59.89 & 58.34 & 77.77 & 69.13 & 71.31 & 0 \\
Ours w/o Interactor & 68.13 & 45.86 & 313.43 & 62.40 & 68.79 & 85.45 & 78.53 & 81.75 & 14.95 \\
Ours w/o Selection & 68.45 & 47.13 & 331.88 & 63.71 & 68.89 & 89.24 & 79.28 & 83.45 & 14.43 \\
Full            & \textbf{70.42} & \textbf{49.08} & \textbf{344.55} & \textbf{65.02} & \textbf{70.07} & \textbf{89.87} & \textbf{83.87} & \textbf{83.94} & \textbf{15.68} \\
\bottomrule
\end{tabular}
\label{tab:ora}
}
\end{table*}

\begin{table}[htb]
\setlength{\tabcolsep}{1.2mm}{
\centering
\caption{Results on NuScenes-MQA.}
\begin{tabular}{ccccc}
\toprule
METHOD & BLUE1 $\uparrow$ & BLUE4 $\uparrow$ & ROUGE\_L $\uparrow$ & ACC $\uparrow$ \\
\midrule
OPT-1.3B        & \textbf{69.8} & 40.4 & 62.6 & 60.4 \\
OPT-1.3B + st   & 64.4 & 36.0 & 47.4 & 63.8 \\
OPT-6.7B        & 67.4 & 38.4 & 62.4 & 61.1 \\
Ours            & 67.4 & \textbf{48.6} & \textbf{67.5} & \textbf{74.4} \\
\bottomrule
\end{tabular}
\label{tab:mqa}
}
\end{table}

\begin{table}[htb]
\setlength{\tabcolsep}{1.2mm}{
\centering
\caption{Results on OmniDrive-NuScenes.}
\begin{tabular}{cccc}
\toprule
METHOD  & CIDEr $\uparrow$ & ROUGE\_L $\uparrow$ & METEOR $\uparrow$\\
\midrule
OmniDrive                   & 68.6  & 32.6 & 38.0  \\
OmniDrive w/o online        & 69.0  & 32.7 & 38.2  \\
Ours                        & \textbf{103.9} & \textbf{38.5} & \textbf{40.1}  \\
\bottomrule
\end{tabular}
\label{tab:omni}
}
\end{table}

\begin{table}[t]
\setlength{\tabcolsep}{1.2mm}{
\centering
\caption{Results on NuInstruct.}
\begin{tabular}{ccccc}
\toprule
METHOD & MAE $\downarrow$ & ACC $\uparrow$ & mAP $\uparrow$ & BLUE4 $\uparrow$ \\
\midrule
Video-LLama & 12.77 & 24.8 & 12.85 & 25.3 \\
BEV-InMLLM & 9.07 & 32.48 & \textbf{20.71} & 35.2 \\
Ours w/o top-k & 5.09 & 43.81 & 13.01 & 46.69 \\
Ours & \textbf{4.33} & \textbf{52.71} & 16.66 & \textbf{69.85} \\
\bottomrule
\end{tabular}
\label{tab:nuinstruct}
}
\end{table}

\begin{table}[t]
\centering
\caption{Results on NuScenes-QA. $\dagger$ represent use Lidar infomation. }
\begin{tabular}{cc}
\toprule
METHOD & ACC $\uparrow$  \\
\midrule
MSMDFusion+MCAN  & \textbf{60.4} \\
CenterPoint+MCAN $\dagger$ & 59.5 \\
OmniDrive & 59.2 \\
Ours            & 58.4 \\
\bottomrule
\end{tabular}
\label{tab:nusqa}
\end{table}

\section{Experiments}
\subsection{Implementation}
We employ EVA-02-L~\cite{EVA02} as the image encoder and a re-trained SparseBEV~\cite{liu2023sparsebev} (excluding future frames and validation set) as the BEV encoder. For the large language model, we utilize LLaMA3-8B~\cite{touvron2023llama}. A 2-layer MLP with ReLU activation functions is used to map feature dimensions. In the selection operation, cosine similarity is used as the similarity metric, and a 2-layer cross-attention is employed in the interaction operation. When selecting $\operatorname{top-k}$ tokens, we set $k=90$ for image features and $k=300$ for BEV features. 

During the single-view and multi-view alignment pre-training stage, we adopt the same strategies as LLava-Next~\cite{liu2024llavanext}, including optimizer, learning rate, and batch size, training for 2 epochs. We use 32 Tesla A100 80G to train 3 days.
In the subsequent task-specific instruction tuning stage, we switch to the AdamW~\cite{loshchilov2017decoupled} optimizer, setting the learning rate to $1 \times 10^{-5}$ and a batch size of 8. To promote training stability and convergence, we implement a cosine annealing learning rate schedule with a warm-up period. For this stage, we use 8 Tesla A100 80G GPUs, and the training is conducted over a period of 8 hours.

\subsection{Matrics}
For caption task such as scene description and risk assessment, we employ commonly used language-based metrics to evaluate word-level sentence similarity, including BLEU, ROUGE\_L, and CIDEr. Notable, for data in NuScenes-MQA~\cite{inoue2024nuscenes} with tagged parts and OmniDrive-NuScenes~\cite{wang2024omnidrive}, we compute the Accuracy metric. 
For grounding tasks, we use the mAP metric to evaluate how well the predicted bounding boxes match the ground truth.
For Visual Question Answering (VQA) tasks conducted on the NuScenes-QA~\cite{qian2024nuscenes} dataset, we differentiate between the types of questions to select appropriate evaluation metrics. 
Questions pertaining to object categories are assessed using the Accuracy metric, which measures the proportion of correctly identified categories. 
In contrast, questions related to spatial attributes such as distance and displacement are evaluated using the Mean Absolute Error (MAE), which quantifies the average magnitude of errors in distance or displacement predictions.

For open-loop driving, we follow standard practices by utilizing the implementation provided by VAD~\cite{jiang2023vad} to evaluate planning within 1, 2, and 3-seconds time horizons. We use two widely accepted metrics to assess performance: the L2 error, calculated by comparing the predicted trajectory of the self-vehicle with the ground truth trajectory at corresponding way-points, and the collision rate, determined by checking for any intersections between the self-vehicle and other entities in the scene.

\subsection{Main Results}
We evaluated our method on NuScenes-MQA ~\cite{inoue2024nuscenes} in Table~\ref{tab:mqa} and OmniDrive-NuScenes~\cite{wang2024omnidrive} in Table~\ref{tab:omni}, and observed significant improvements across various metrics. 
Specifically, in the NuScenes-MQA dataset, ACC measures the average accuracy of yes/no questions, classification tasks, and counting tasks under correct category conditions. Our approach achieves a 10.6\% improvement over the previous state-of-the-art (SoTA) methods. In the OmniDrive-NuScenes dataset, we evaluated caption-based metrics, demonstrating a 51.4\% improvement in CIDEr.

As shown in Table~\ref{tab:ora}, for the object-level risk assessment dataset, our evaluation is divided into three components: the language score evaluate the quality of risk explanation. Accuracy measures the precision of risk information (categories, levels, objects, and presence), where categories, levels, and objects are evaluated only when the exists is correctly. mAP evaluates the localization of the maximum risk targets. We evaluated both the baseline model and our model with different added modules, and our complete model achieved optimal performance.

Furthermore, we utilized the NuInstruct~\cite{ding2024holistic} dataset to assess our method’s capability to handle multi-view information, where we also observed notable improvements across all metrics. The results are shown in Table~\ref{tab:nuinstruct}.
Specifically, we compute the average MAE for distance, speed, count and motion. We calculate the average ACC for loset object, status and is in the same road. 
For the risk tasks, we employ the mAP metric, and for the reasoning tasks, we use BLUE4. 
Our approach achieves SoTA performance in MAE, ACC, and BLUE4, with a 20.23\% improvement in the ACC metric and 98.44\% improvement in the BLUE4 metric. We also achieve comparable performance in the mAP metric.

For the VQA task on NuScenes-QA shown in Table~\ref{tab:nusqa}, we also have achieved comparable performance.
These experimental results robustly demonstrate the effectiveness of our proposed method.

\subsection{Ablation Studies}

As shown in Table~\ref{tab:ablation}, we conducted ablation studies on the NuScenes-QA~\cite{qian2024nuscenes} and OmniDrive-NuScenes~\cite{wang2024omnidrive} datasets to validate the effectiveness of our proposed module. 
And in Table~\ref{tab:ora}, we present the impact of different modules on reasoning abilities under perception-limited conditions.


Experimental results demonstrate the critical role of our training strategy and dataset in enhancing the grounding task performance. 
When we incorporated BEV representations, there was a noticeable improvement in the grounding task’s performance. 
However, this addition had a relatively minor impact on captioning tasks, indicating that BEV benefits are more pronounced in grounding than in captioning.
Moreover, integrating the interactor component without the $\operatorname{top-k}$ operation yielded substantial improvements across various evaluation metrics. 
This enhancement is attributed to the effective integration of instruction-guided information, which considerably boosts performance. 
The $\operatorname{top-k}$ operation, which aggregates information more efficiently, further optimizes the system’s capabilities. 
Its inclusion facilitates a more nuanced understanding by the Large Language Model (LLM), leading to the best overall performance of our complete model.


\begin{table}[htb]
\centering
\setlength{\tabcolsep}{1.2mm}{
\caption{Ablation study on OmniDrive-NuScenes, NuScenes-MQA and NuInstruct datasets. TS represents our three stage training strategy.}
\begin{tabular}{lcccc}
\toprule
Model & ACC $\uparrow$ & CIDEr $\uparrow$ & BLUE4 $\uparrow$ & mAP $\downarrow$ \\
\midrule
            Base            & 64.2 & 70.2 & 36.4 & 0 \\
\;          + TS            & 71.9 & 92.1 & 36.8 & 10.41 \\
\;\;        + BEV           & 73.3 & 93.6 & 36.7 & 11.66 \\
\;\;\;      + Interactor    & 74.1 & 101.2 & 41.69 & 13.01 \\
\;\;\;\;    + Selection     & \textbf{74.4} & \textbf{103.9} & \textbf{49.08} & \textbf{16.66} \\
\bottomrule
\end{tabular}
\label{tab:ablation}
}
\end{table}

\begin{table*}[htb]
\centering
\caption{Comparsions on the open-loos planning. For a fair comparison, we refer to the reproduced results in BEV-Planner. The bold numbers represent the highest accuracy. The optimal results are highlighted in bold.}
\begin{tabular}{c|c|cc|cccc|cccc}
\toprule
\multirow{2}{*}{METHOD} & 
HIGH LEVEL &
\multicolumn{2}{c|}{EGO STATUS} & 
\multicolumn{4}{c|}{L2 (m) $\downarrow$ } & 
\multicolumn{4}{c}{COLLISION (\%) $\downarrow$ } \\
& COMMEND & BEV & Planner & 1s & 2s & 3s & AVG & 1s & 2s & 3s & AVG \\
\midrule 
\multirow{2}{*}{UniAD} 
     & \cmark & \xmark & \xmark & 0.59 & 1.01 & 1.48 & 1.03 & 0.16 & 0.51 & 1.64 & 0.77 \\
     & \cmark & \cmark & \cmark & 0.20 & 0.42 & 0.75 & 0.46 & 0.02 & 0.25 & 0.84 & 0.37 \\
\midrule
\multirow{2}{*}{VAD-Base} 
     & \cmark & \xmark & \xmark & 0.69 & 1.22 & 1.83 & 1.25 & 0.06 & 0.68 & 2.52 & 1.09 \\
     & \cmark & \cmark & \cmark & 0.17 & 0.34 & 0.60 & 0.37 & 0.04 & 0.27 & 0.67 & 0.33 \\
\midrule
\multirow{2}{*}{BEV-Planner} 
     & \cmark & \xmark & \xmark & 0.30 & 0.52 & 0.83 & 0.55 & 0.10 & 0.37 & 1.30 & 0.59 \\
     & \cmark & \cmark & \cmark & 0.16 & 0.32 & 0.57 & 0.35 & 0.00 & 0.29 & 0.73 & 0.34 \\
\midrule
\multirow{3}{*}{OmniDrive} 
     & \xmark & \xmark & \xmark & 1.15 & 1.96 & 2.84 & 1.98 & 0.80 & 3.12 & 7.46 & 3.79 \\
     & \cmark & \xmark & \xmark & 0.40 & 0.80 & 1.32 & 0.84 & 0.04 & 0.46 & 2.32 & 0.94 \\
     & \cmark & \cmark & \cmark & 0.14 & \textbf{0.29} & 0.55 & 0.33 & \textbf{0.00} & \textbf{0.13} & 0.78 & 0.30 \\
\midrule
\multirow{3}{*}{Ours} 
     & \xmark & \xmark & \xmark & 0.3 & 0.65 & 1.14 & 0.7 & 0.14 & 0.49 & 1.03 & 0.55 \\
     & \cmark & \xmark & \xmark & 0.29 & 0.6 & 1.03 & 0.64 & 0.1 & 0.2 & 0.51 & 0.27 \\
     & \cmark & \cmark & \cmark & \textbf{0.14} & 0.3 & \textbf{0.55} & \textbf{0.33} & 0.07 & 0.14 & \textbf{0.32} & \textbf{0.18} \\
\bottomrule
\end{tabular}
\label{tab:openloop}
\end{table*}

\begin{figure}[t]
    \centering
    \includegraphics[width=\linewidth]{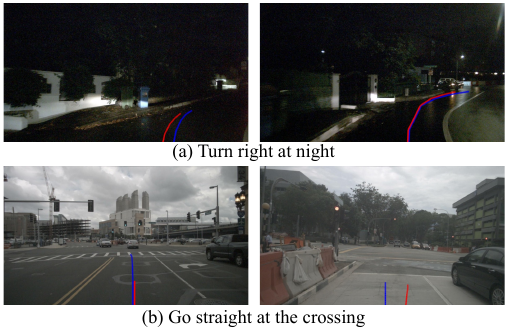}
    \caption{\textbf{Qualitative results with planning.} The red line represents the ground truth path, while the blue line indicates the path predicted by our method. These results were obtained without ego status.}
    \label{fig:viz}
\end{figure}

\begin{table}[t]
\centering
\setlength{\tabcolsep}{1.2mm}{
\caption{Parameter analyze on OmniDrive-NuScenes, NuScenes-QA and NuInstruct datasets}
\begin{tabular}{cccccc}
\toprule
$k$ & BLUE4 $\uparrow$ & CIDEr $\uparrow$ & ROUGE\_L $\uparrow$ & ACC $\uparrow$ & mAP $\uparrow$ \\
\midrule
30  & 14.2 & \textbf{105.3} & 38.5 & 57.3 & 13.01 \\
60  & 13.1 & 98.9 & 38.3 & 54.8 & 11.6 \\
90  & \textbf{16.9} & 103.9 & 38.5 & \textbf{57.4} & \textbf{16.66} \\
120 & 15.3 & 102.2 & \textbf{38.6} & 56.8 & 12.5 \\
\bottomrule
\end{tabular}
\label{tab:params}
}
\end{table}

\paragraph{Parameter Analyzation.}
We analyzed the impact of the value of  $k$  on model performance. As shown in Table~\ref{tab:params}, the optimal performance was achieved when  $k=90$ . We attribute this to the redundancy present in the data. 
There is a significant amount of information in the visual tokens that is irrelevant to the instructions. 
We utilized a selection module to extract the $k$ most relevant visual tokens. 
However, a value of $k$ that is too small leads to the loss of key information, while a value that is too large fails to mitigate redundancy. 

\subsection{Open-loop Planning}
We compare our method with previous SoTA approaches in Table ~\ref{tab:openloop}. 
We adopt a distinct encoding scheme for ego status: firstly, we convert all units from meters to centimeters and round to the nearest integer to facilitate tokenization by the language model. Subsequently, ego status is input to the large language model in a linguistic format 
(\textit{e.g.} ``Given the ego status: lateral velocity is 0 $cm/s$; longitudinal velocity is 418 $cm/s$; lateral acceleration is 5 $cm/s^2$; longitudinal acceleration is 93 $cm/s^2$; The ego car will TURN LEFT. Output planning results.'').


As described in BEV-Planner~\cite{li2024ego}, encoding ego status can significantly enhance the performance of all methods. Therefore, we conducted experiments focusing on both ego status and high-level commands. 
Significantly, for our approach, encoding ego status substantially improve planning performance, whereas high-level commands offer limited improvements in planning performance. 
Upon analysis, we consider that in the nuScenes~\cite{nuscenes2019} scenario, the driving behavior (high-level commends) choices available in most scenarios are unique, and our model is capable of fully perceiving the current scene to make current plans.

Our method achieves a new SoTA performance in collision rate and reaches comparable in L2 error. 
Additionally, our method attains SoTA performance even without incorporating high-level commands and ego status. 
When employing high-level commands but omitting ego status, our method also achieves SoTA performance in collision rate and demonstrates comparable results in L2 error.



\subsection{Qualitative Results}
We visualized the planning results of open-loop driving without ego status to better understand the effectiveness of our approach. 
As shown in Figure~\ref{fig:viz}, our method, while producing higher L2 errors after the training phase, demonstrates notable improvements in the quality of the planning paths generated. 
For example, as shown in Figure~\ref{fig:viz} (a), a larger steering angle enables the ego vehicle to complete turns more quickly. In Figure~\ref{fig:viz} (b), in the scenario with a traffic light at an intersection, the model decelerates and stops when the light is red. However, when the light is green, the model accelerates through the intersection, which differs from the ground truth.

The qualitative results reveals that our approach consistently generates paths that are more reasonable and practical compared to previous methods. This enhanced path generation capability is not merely a theoretical improvement but translates into significant practical benefits. 
Overall, while the L2 error did not show a significant decrease, the qualitative improvements in path planning and the substantial reduction in collision occurrences underscore the effectiveness and practicality of our method in open-loop driving scenarios.

\section{Conclusion}
In this paper, we propose a framework to integrate world-knowledge and perception ability. 
By combining a instruction-guided interaction module, our approach effectively fuses multi-view video data with natural language instructions, leading to enriched visual representations. 
Then, we collected and refined a large-scale multi-modal dataset that includes 2 million natural language QA pairs, 1.7 million grounding data. 
The risk assessment data validates the performance of our approach under perception-limited conditions.
Extensive experiments across tasks such as VQA, open-loop driving, and detection demonstrate the effectiveness, comprehensiveness, and generalization of our approach.

\bibliography{aaai25}

\begin{thebibliography}{47}
\providecommand{\natexlab}[1]{#1}

\bibitem[{Achiam et~al.(2023)Achiam, Adler, Agarwal, Ahmad, Akkaya, Aleman, Almeida, Altenschmidt, Altman, Anadkat et~al.}]{achiam2023gpt}
Achiam, J.; Adler, S.; Agarwal, S.; Ahmad, L.; Akkaya, I.; Aleman, F.~L.; Almeida, D.; Altenschmidt, J.; Altman, S.; Anadkat, S.; et~al. 2023.
\newblock Gpt-4 technical report.
\newblock \emph{arXiv preprint arXiv:2303.08774}.

\bibitem[{Alayrac et~al.(2022)Alayrac, Donahue, Luc, Miech, Barr, Hasson, Lenc, Mensch, Millican, Reynolds et~al.}]{alayrac2022flamingo}
Alayrac, J.-B.; Donahue, J.; Luc, P.; Miech, A.; Barr, I.; Hasson, Y.; Lenc, K.; Mensch, A.; Millican, K.; Reynolds, M.; et~al. 2022.
\newblock Flamingo: a visual language model for few-shot learning.
\newblock \emph{Advances in neural information processing systems}, 35: 23716--23736.

\bibitem[{Awadalla et~al.(2023)Awadalla, Gao, Gardner, Hessel, Hanafy, Zhu, Marathe, Bitton, Gadre, Sagawa et~al.}]{awadalla2023openflamingo}
Awadalla, A.; Gao, I.; Gardner, J.; Hessel, J.; Hanafy, Y.; Zhu, W.; Marathe, K.; Bitton, Y.; Gadre, S.; Sagawa, S.; et~al. 2023.
\newblock Openflamingo: An open-source framework for training large autoregressive vision-language models.
\newblock \emph{arXiv preprint arXiv:2308.01390}.

\bibitem[{Bai et~al.(2023)Bai, Bai, Yang, Wang, Tan, Wang, Lin, Zhou, and Zhou}]{bai2023qwen}
Bai, J.; Bai, S.; Yang, S.; Wang, S.; Tan, S.; Wang, P.; Lin, J.; Zhou, C.; and Zhou, J. 2023.
\newblock Qwen-vl: A frontier large vision-language model with versatile abilities.
\newblock \emph{arXiv preprint arXiv:2308.12966}.

\bibitem[{Bai et~al.(2024)Bai, Wu, Liu, Jia, Mao, Zhang, Zhao, Shen, Wei, Wang et~al.}]{bai20243d}
Bai, Y.; Wu, D.; Liu, Y.; Jia, F.; Mao, W.; Zhang, Z.; Zhao, Y.; Shen, J.; Wei, X.; Wang, T.; et~al. 2024.
\newblock Is a 3D-Tokenized LLM the Key to Reliable Autonomous Driving?
\newblock \emph{arXiv preprint arXiv:2405.18361}.

\bibitem[{Brohan et~al.(2023)Brohan, Brown, Carbajal, Chebotar, Chen, Choromanski, Ding, Driess, Dubey, Finn et~al.}]{brohan2023rt}
Brohan, A.; Brown, N.; Carbajal, J.; Chebotar, Y.; Chen, X.; Choromanski, K.; Ding, T.; Driess, D.; Dubey, A.; Finn, C.; et~al. 2023.
\newblock Rt-2: Vision-language-action models transfer web knowledge to robotic control.
\newblock \emph{arXiv preprint arXiv:2307.15818}.

\bibitem[{Caesar et~al.(2019)Caesar, Bankiti, Lang, Vora, Liong, Xu, Krishnan, Pan, Baldan, and Beijbom}]{nuscenes2019}
Caesar, H.; Bankiti, V.; Lang, A.~H.; Vora, S.; Liong, V.~E.; Xu, Q.; Krishnan, A.; Pan, Y.; Baldan, G.; and Beijbom, O. 2019.
\newblock nuScenes: A multimodal dataset for autonomous driving.
\newblock \emph{arXiv preprint arXiv:1903.11027}.

\bibitem[{Chen et~al.(2024{\natexlab{a}})Chen, Sinavski, Hünermann, Karnsund, Willmott, Birch, Maund, and Shotton}]{chen2024drivingwithllms}
Chen, L.; Sinavski, O.; Hünermann, J.; Karnsund, A.; Willmott, A.~J.; Birch, D.; Maund, D.; and Shotton, J. 2024{\natexlab{a}}.
\newblock Driving with LLMs: Fusing Object-Level Vector Modality for Explainable Autonomous Driving.
\newblock In \emph{2024 IEEE International Conference on Robotics and Automation (ICRA)}.

\bibitem[{Chen et~al.(2022)Chen, Wang, Changpinyo, Piergiovanni, Padlewski, Salz, Goodman, Grycner, Mustafa, Beyer et~al.}]{chen2022pali}
Chen, X.; Wang, X.; Changpinyo, S.; Piergiovanni, A.; Padlewski, P.; Salz, D.; Goodman, S.; Grycner, A.; Mustafa, B.; Beyer, L.; et~al. 2022.
\newblock Pali: A jointly-scaled multilingual language-image model.
\newblock \emph{arXiv preprint arXiv:2209.06794}.

\bibitem[{Chen et~al.(2024{\natexlab{b}})Chen, Wu, Wang, Su, Chen, Xing, Zhong, Zhang, Zhu, Lu et~al.}]{chen2024internvl}
Chen, Z.; Wu, J.; Wang, W.; Su, W.; Chen, G.; Xing, S.; Zhong, M.; Zhang, Q.; Zhu, X.; Lu, L.; et~al. 2024{\natexlab{b}}.
\newblock Internvl: Scaling up vision foundation models and aligning for generic visual-linguistic tasks.
\newblock In \emph{Proceedings of the IEEE/CVF Conference on Computer Vision and Pattern Recognition}, 24185--24198.

\bibitem[{Cohen(1997)}]{cohen1997align}
Cohen, G.~H. 1997.
\newblock ALIGN: a program to superimpose protein coordinates, accounting for insertions and deletions.
\newblock \emph{Journal of applied crystallography}, 30(6): 1160--1161.

\bibitem[{Cui et~al.(2023)Cui, Huang, Zhong, Liu, Wang, Sun, Li, Wang, and Khajepour}]{cui2023drivellm}
Cui, Y.; Huang, S.; Zhong, J.; Liu, Z.; Wang, Y.; Sun, C.; Li, B.; Wang, X.; and Khajepour, A. 2023.
\newblock Drivellm: Charting the path toward full autonomous driving with large language models.
\newblock \emph{IEEE Transactions on Intelligent Vehicles}.

\bibitem[{Dewangan et~al.(2023)Dewangan, Choudhary, Chandhok, Priyadarshan, Jain, Singh, Srivastava, Jatavallabhula, and Krishna}]{dewangan2023talk2bev}
Dewangan, V.; Choudhary, T.; Chandhok, S.; Priyadarshan, S.; Jain, A.; Singh, A.~K.; Srivastava, S.; Jatavallabhula, K.~M.; and Krishna, K.~M. 2023.
\newblock Talk2BEV: Language-enhanced Bird's-eye View Maps for Autonomous Driving.
\newblock \emph{arXiv preprint arXiv:2310.02251}.

\bibitem[{Ding et~al.(2024)Ding, Han, Xu, Liang, Zhang, and Li}]{ding2024holistic}
Ding, X.; Han, J.; Xu, H.; Liang, X.; Zhang, W.; and Li, X. 2024.
\newblock Holistic Autonomous Driving Understanding by Bird's-Eye-View Injected Multi-Modal Large Models.
\newblock In \emph{Proceedings of the IEEE/CVF Conference on Computer Vision and Pattern Recognition}, 13668--13677.

\bibitem[{Ding et~al.(2023)Ding, Han, Xu, Zhang, and Li}]{ding2023hilm}
Ding, X.; Han, J.; Xu, H.; Zhang, W.; and Li, X. 2023.
\newblock Hilm-d: Towards high-resolution understanding in multimodal large language models for autonomous driving.
\newblock \emph{arXiv preprint arXiv:2309.05186}.

\bibitem[{Dosovitskiy et~al.(2017)Dosovitskiy, Ros, Codevilla, Lopez, and Koltun}]{Dosovitskiy17}
Dosovitskiy, A.; Ros, G.; Codevilla, F.; Lopez, A.; and Koltun, V. 2017.
\newblock {CARLA}: {An} Open Urban Driving Simulator.
\newblock In \emph{Proceedings of the 1st Annual Conference on Robot Learning}, 1--16.

\bibitem[{Driess et~al.(2023)Driess, Xia, Sajjadi, Lynch, Chowdhery, Ichter, Wahid, Tompson, Vuong, Yu et~al.}]{driess2023palm}
Driess, D.; Xia, F.; Sajjadi, M.~S.; Lynch, C.; Chowdhery, A.; Ichter, B.; Wahid, A.; Tompson, J.; Vuong, Q.; Yu, T.; et~al. 2023.
\newblock Palm-e: An embodied multimodal language model.
\newblock \emph{arXiv preprint arXiv:2303.03378}.

\bibitem[{Fang et~al.(2023)Fang, Sun, Wang, Huang, Wang, and Cao}]{EVA02}
Fang, Y.; Sun, Q.; Wang, X.; Huang, T.; Wang, X.; and Cao, Y. 2023.
\newblock EVA-02: A Visual Representation for Neon Genesis.
\newblock \emph{arXiv preprint arXiv:2303.11331}.

\bibitem[{Fu et~al.(2024)Fu, Li, Wen, Dou, Cai, Shi, and Qiao}]{fu2024drive}
Fu, D.; Li, X.; Wen, L.; Dou, M.; Cai, P.; Shi, B.; and Qiao, Y. 2024.
\newblock Drive like a human: Rethinking autonomous driving with large language models.
\newblock In \emph{Proceedings of the IEEE/CVF Winter Conference on Applications of Computer Vision}, 910--919.

\bibitem[{GLM et~al.(2024)GLM, Zeng, Xu, Wang, Zhang, Yin, Rojas, Feng, Zhao, Lai, Yu, Wang, Sun, Zhang, Cheng, Gui, Tang, Zhang, Li, Zhao, Wu, Zhong, Liu, Huang, Zhang, Zheng, Lu, Duan, Zhang, Cao, Yang, Tam, Zhao, Liu, Xia, Zhang, Gu, Lv, Liu, Liu, Yang, Song, Zhang, An, Xu, Niu, Yang, Li, Bai, Dong, Qi, Wang, Yang, Du, Hou, and Wang}]{glm2024chatglm}
GLM, T.; Zeng, A.; Xu, B.; Wang, B.; Zhang, C.; Yin, D.; Rojas, D.; Feng, G.; Zhao, H.; Lai, H.; Yu, H.; Wang, H.; Sun, J.; Zhang, J.; Cheng, J.; Gui, J.; Tang, J.; Zhang, J.; Li, J.; Zhao, L.; Wu, L.; Zhong, L.; Liu, M.; Huang, M.; Zhang, P.; Zheng, Q.; Lu, R.; Duan, S.; Zhang, S.; Cao, S.; Yang, S.; Tam, W.~L.; Zhao, W.; Liu, X.; Xia, X.; Zhang, X.; Gu, X.; Lv, X.; Liu, X.; Liu, X.; Yang, X.; Song, X.; Zhang, X.; An, Y.; Xu, Y.; Niu, Y.; Yang, Y.; Li, Y.; Bai, Y.; Dong, Y.; Qi, Z.; Wang, Z.; Yang, Z.; Du, Z.; Hou, Z.; and Wang, Z. 2024.
\newblock ChatGLM: A Family of Large Language Models from GLM-130B to GLM-4 All Tools.
\newblock arXiv:2406.12793.

\bibitem[{H.~Caesar(2021)}]{nuplan}
H.~Caesar, K. T. e.~a., J.~Kabzan. 2021.
\newblock NuPlan: A closed-loop ML-based planning benchmark for autonomous vehicles.
\newblock In \emph{CVPR ADP3 workshop}.

\bibitem[{Inoue et~al.(2024)Inoue, Yada, Tanahashi, and Yamaguchi}]{inoue2024nuscenes}
Inoue, Y.; Yada, Y.; Tanahashi, K.; and Yamaguchi, Y. 2024.
\newblock Nuscenes-mqa: Integrated evaluation of captions and qa for autonomous driving datasets using markup annotations.
\newblock In \emph{Proceedings of the IEEE/CVF Winter Conference on Applications of Computer Vision}, 930--938.

\bibitem[{Jiang et~al.(2023)Jiang, Chen, Xu, Liao, Chen, Zhou, Zhang, Liu, Huang, and Wang}]{jiang2023vad}
Jiang, B.; Chen, S.; Xu, Q.; Liao, B.; Chen, J.; Zhou, H.; Zhang, Q.; Liu, W.; Huang, C.; and Wang, X. 2023.
\newblock Vad: Vectorized scene representation for efficient autonomous driving.
\newblock In \emph{Proceedings of the IEEE/CVF International Conference on Computer Vision}, 8340--8350.

\bibitem[{Jiang et~al.(2024)Jiang, He, Zeng, Wei, Ku, Liu, and Chen}]{jiang2024mantis}
Jiang, D.; He, X.; Zeng, H.; Wei, C.; Ku, M.; Liu, Q.; and Chen, W. 2024.
\newblock Mantis: Interleaved multi-image instruction tuning.
\newblock \emph{arXiv preprint arXiv:2405.01483}.

\bibitem[{Lauren{\c{c}}on et~al.(2024)Lauren{\c{c}}on, Tronchon, Cord, and Sanh}]{laurenccon2024matters}
Lauren{\c{c}}on, H.; Tronchon, L.; Cord, M.; and Sanh, V. 2024.
\newblock What matters when building vision-language models?
\newblock \emph{arXiv preprint arXiv:2405.02246}.

\bibitem[{Li et~al.(2023)Li, Li, Savarese, and Hoi}]{li2023blip}
Li, J.; Li, D.; Savarese, S.; and Hoi, S. 2023.
\newblock Blip-2: Bootstrapping language-image pre-training with frozen image encoders and large language models.
\newblock In \emph{International conference on machine learning}, 19730--19742. PMLR.

\bibitem[{Li et~al.(2024)Li, Yu, Lan, Li, Kautz, Lu, and Alvarez}]{li2024ego}
Li, Z.; Yu, Z.; Lan, S.; Li, J.; Kautz, J.; Lu, T.; and Alvarez, J.~M. 2024.
\newblock Is ego status all you need for open-loop end-to-end autonomous driving?
\newblock In \emph{Proceedings of the IEEE/CVF Conference on Computer Vision and Pattern Recognition}, 14864--14873.

\bibitem[{Lin et~al.(2024)Lin, Yin, Ping, Molchanov, Shoeybi, and Han}]{lin2024vila}
Lin, J.; Yin, H.; Ping, W.; Molchanov, P.; Shoeybi, M.; and Han, S. 2024.
\newblock Vila: On pre-training for visual language models.
\newblock In \emph{Proceedings of the IEEE/CVF Conference on Computer Vision and Pattern Recognition}, 26689--26699.

\bibitem[{Liu et~al.(2024{\natexlab{a}})Liu, Li, Li, Li, Zhang, Shen, and Lee}]{liu2024llavanext}
Liu, H.; Li, C.; Li, Y.; Li, B.; Zhang, Y.; Shen, S.; and Lee, Y.~J. 2024{\natexlab{a}}.
\newblock LLaVA-NeXT: Improved reasoning, OCR, and world knowledge.

\bibitem[{Liu et~al.(2024{\natexlab{b}})Liu, Li, Wu, and Lee}]{liu2024visual}
Liu, H.; Li, C.; Wu, Q.; and Lee, Y.~J. 2024{\natexlab{b}}.
\newblock Visual instruction tuning.
\newblock \emph{Advances in neural information processing systems}, 36.

\bibitem[{Liu et~al.(2023{\natexlab{a}})Liu, Teng, Lu, Wang, and Wang}]{liu2023sparsebev}
Liu, H.; Teng, Y.; Lu, T.; Wang, H.; and Wang, L. 2023{\natexlab{a}}.
\newblock Sparsebev: High-performance sparse 3d object detection from multi-camera videos.
\newblock In \emph{Proceedings of the IEEE/CVF International Conference on Computer Vision}, 18580--18590.

\bibitem[{Liu et~al.(2023{\natexlab{b}})Liu, Zeng, Ren, Li, Zhang, Yang, Li, Yang, Su, Zhu et~al.}]{liu2023grounding}
Liu, S.; Zeng, Z.; Ren, T.; Li, F.; Zhang, H.; Yang, J.; Li, C.; Yang, J.; Su, H.; Zhu, J.; et~al. 2023{\natexlab{b}}.
\newblock Grounding dino: Marrying dino with grounded pre-training for open-set object detection.
\newblock \emph{arXiv preprint arXiv:2303.05499}.

\bibitem[{Loshchilov and Hutter(2017)}]{loshchilov2017decoupled}
Loshchilov, I.; and Hutter, F. 2017.
\newblock Decoupled weight decay regularization.
\newblock \emph{arXiv preprint arXiv:1711.05101}.

\bibitem[{Ma et~al.(2023)Ma, Cao, Sun, Pavone, and Xiao}]{ma2023dolphins}
Ma, Y.; Cao, Y.; Sun, J.; Pavone, M.; and Xiao, C. 2023.
\newblock Dolphins: Multimodal Language Model for Driving.
\newblock \emph{arXiv prepreint arXiv:2312.00438}.

\bibitem[{Mei et~al.(2024)Mei, Ma, Yang, Wen, Cai, Li, Fu, Zhang, Cai, Dou et~al.}]{mei2024continuously}
Mei, J.; Ma, Y.; Yang, X.; Wen, L.; Cai, X.; Li, X.; Fu, D.; Zhang, B.; Cai, P.; Dou, M.; et~al. 2024.
\newblock Continuously Learning, Adapting, and Improving: A Dual-Process Approach to Autonomous Driving.
\newblock \emph{arXiv preprint arXiv:2405.15324}.

\bibitem[{Qian et~al.(2024)Qian, Chen, Zhuo, Jiao, and Jiang}]{qian2024nuscenes}
Qian, T.; Chen, J.; Zhuo, L.; Jiao, Y.; and Jiang, Y.-G. 2024.
\newblock Nuscenes-qa: A multi-modal visual question answering benchmark for autonomous driving scenario.
\newblock In \emph{Proceedings of the AAAI Conference on Artificial Intelligence}, volume~38, 4542--4550.

\bibitem[{Radford et~al.(2021)Radford, Kim, Hallacy, Ramesh, Goh, Agarwal, Sastry, Askell, Mishkin, Clark et~al.}]{radford2021learning}
Radford, A.; Kim, J.~W.; Hallacy, C.; Ramesh, A.; Goh, G.; Agarwal, S.; Sastry, G.; Askell, A.; Mishkin, P.; Clark, J.; et~al. 2021.
\newblock Learning transferable visual models from natural language supervision.
\newblock In \emph{International conference on machine learning}, 8748--8763. PMLR.

\bibitem[{Sima et~al.(2023)Sima, Renz, Chitta, Chen, Zhang, Xie, Luo, Geiger, and Li}]{sima2023drivelm}
Sima, C.; Renz, K.; Chitta, K.; Chen, L.; Zhang, H.; Xie, C.; Luo, P.; Geiger, A.; and Li, H. 2023.
\newblock DriveLM: Driving with Graph Visual Question Answering.
\newblock \emph{arXiv preprint arXiv:2312.14150}.

\bibitem[{Sun et~al.(2024)Sun, Cui, Zhang, Zhang, Yu, Wang, Rao, Liu, Huang, and Wang}]{sun2024generative}
Sun, Q.; Cui, Y.; Zhang, X.; Zhang, F.; Yu, Q.; Wang, Y.; Rao, Y.; Liu, J.; Huang, T.; and Wang, X. 2024.
\newblock Generative multimodal models are in-context learners.
\newblock In \emph{Proceedings of the IEEE/CVF Conference on Computer Vision and Pattern Recognition}, 14398--14409.

\bibitem[{Tian et~al.(2024{\natexlab{a}})Tian, Li, Weng, Chen, Schmerling, Wang, Ivanovic, and Pavone}]{tian2024tokenize}
Tian, R.; Li, B.; Weng, X.; Chen, Y.; Schmerling, E.; Wang, Y.; Ivanovic, B.; and Pavone, M. 2024{\natexlab{a}}.
\newblock Tokenize the World into Object-level Knowledge to Address Long-tail Events in Autonomous Driving.
\newblock \emph{arXiv preprint arXiv:2407.00959}.

\bibitem[{Tian et~al.(2024{\natexlab{b}})Tian, Gu, Li, Liu, Zhao, Wang, Zhan, Jia, Lang, and Zhao}]{DriveVLM}
Tian, X.; Gu, J.; Li, B.; Liu, Y.; Zhao, Z.; Wang, Y.; Zhan, K.; Jia, P.; Lang, X.; and Zhao, H. 2024{\natexlab{b}}.
\newblock DriveVLM: The Convergence of Autonomous Driving and Large Vision-Language Models.
\newblock \emph{arXiv preprint arXiv:2402.12289}.

\bibitem[{Touvron et~al.(2023)Touvron, Martin, Stone, Albert, Almahairi, Babaei, Bashlykov, Batra, Bhargava, Bhosale et~al.}]{touvron2023llama}
Touvron, H.; Martin, L.; Stone, K.; Albert, P.; Almahairi, A.; Babaei, Y.; Bashlykov, N.; Batra, S.; Bhargava, P.; Bhosale, S.; et~al. 2023.
\newblock Llama 2: Open foundation and fine-tuned chat models.
\newblock \emph{arXiv preprint arXiv:2307.09288}.

\bibitem[{Wang et~al.(2024{\natexlab{a}})Wang, Yu, Jiang, Lan, Shi, Chang, Kautz, Li, and Alvarez}]{wang2024omnidrive}
Wang, S.; Yu, Z.; Jiang, X.; Lan, S.; Shi, M.; Chang, N.; Kautz, J.; Li, Y.; and Alvarez, J.~M. 2024{\natexlab{a}}.
\newblock OmniDrive: A Holistic LLM-Agent Framework for Autonomous Driving with 3D Perception, Reasoning and Planning.
\newblock \emph{arXiv preprint arXiv:2405.01533}.

\bibitem[{Wang et~al.(2024{\natexlab{b}})Wang, Xie, Chu, Li, and Luo}]{wang2024drivecot}
Wang, T.; Xie, E.; Chu, R.; Li, Z.; and Luo, P. 2024{\natexlab{b}}.
\newblock Drivecot: Integrating chain-of-thought reasoning with end-to-end driving.
\newblock \emph{arXiv preprint arXiv:2403.16996}.

\bibitem[{Wang et~al.(2023)Wang, Xie, Hu, Zou, Fan, Tong, Wen, Wu, Deng, Li et~al.}]{wang2023drivemlm}
Wang, W.; Xie, J.; Hu, C.; Zou, H.; Fan, J.; Tong, W.; Wen, Y.; Wu, S.; Deng, H.; Li, Z.; et~al. 2023.
\newblock Drivemlm: Aligning multi-modal large language models with behavioral planning states for autonomous driving.
\newblock \emph{arXiv preprint arXiv:2312.09245}.

\bibitem[{Wen et~al.(2023)Wen, Fu, Li, Cai, Ma, Cai, Dou, Shi, He, and Qiao}]{wen2023dilu}
Wen, L.; Fu, D.; Li, X.; Cai, X.; Ma, T.; Cai, P.; Dou, M.; Shi, B.; He, L.; and Qiao, Y. 2023.
\newblock Dilu: A knowledge-driven approach to autonomous driving with large language models.
\newblock \emph{arXiv preprint arXiv:2309.16292}.

\bibitem[{Yu et~al.(2024)Yu, Wang, Tu, Cao, Zhang-Li, Lv, Peng, Yao, Zhang, Li et~al.}]{yukola}
Yu, J.; Wang, X.; Tu, S.; Cao, S.; Zhang-Li, D.; Lv, X.; Peng, H.; Yao, Z.; Zhang, X.; Li, H.; et~al. 2024.
\newblock KoLA: Carefully Benchmarking World Knowledge of Large Language Models.
\newblock In \emph{The Twelfth International Conference on Learning Representations}.

\end{thebibliography}

\clearpage
\onecolumn
\section*{\centering{Supplementary Material}}

\maketitle

\section{Dataset Details}
\paragraph{Data Generation.}
Our data generation process is composed of two main steps:
1. As shown in Table~\ref{tab:extract_risk}, we start by using GPT-4o to identify and generate all potential risks associated with a given object. This step ensures a comprehensive enumeration of possible risks, which serves as the foundation for subsequent analysis.
2. In the second step, as illustrated in Table~\ref{tab:extract_qa}, GPT-4o-mini is employed to systematically organize the identified risks into a question-answer pairs. This structure is crucial for the clear and efficient communication of risk-related information.

During the first step, we specifically extract the locations of high-risk objects, which are then utilized as the primary data for the task of risk target localization. All grounding bounding boxes identified in this process are normalized to fit within a coordinate range of 0-999, ensuring consistency across the dataset.

\paragraph{Data Refine.}

We undertook a comprehensive refinement of four widely-used open-source datasets, including NuScenes-QA~\cite{qian2024nuscenes}, NuScenes-MQA~\cite{inoue2024nuscenes}, OmniDrive-NuScenes~\cite{wang2024omnidrive}, and NuInstruct~\cite{ding2024holistic}. Our refinement process was meticulous and systematic, ensuring the datasets were optimized for accuracy and consistency. This process involved several key steps:

\begin{enumerate}
    \item We began by distinguishing between short answers and long answers in the datasets to ensure clear categorization and reduce ambiguity in data interpretation.
    \item We identified and removed inaccurate bounding boxes, which were crucial for enhancing the precision of object localization and reducing potential errors in downstream tasks.
    \item We converted all decimal values to integers, either through unit conversion or rounding, to maintain uniformity in numerical data representation.
    \item Standardization was also a major focus, where we standardized all tags across the datasets, such as $\texttt{<ref>, <box>, <|camera\_front|>}$ etc, among others, to facilitate easier data parsing and integration across different modules.
    \item We unified the representation of all trajectory data, creating a consistent framework that supports seamless analysis and model development across various datasets.
\end{enumerate}

This expanded version adds more detail to each step, making it clearer to understand the importance and impact of each aspect of the refinement process.

\begin{table*}[htb]
{
\centering
\caption{Comprehensive experimental results. FT means fine-tune. PT1 means single-view pretrain. PT2 means multi-view pretrain. ITA means interactor module. SEL means selection operation.}
\begin{tabular}{c|l|cccccc}
\toprule
DATASET & METHOD & BLUE1 $\uparrow$ & BLUE2 $\uparrow$ & BLUE3 $\uparrow$ & BLUE4 $\uparrow$ & CIDEr $\uparrow$ & ROUGE\_L $\uparrow$ \\

\midrule
\multirow{5}{*}{OmniDrive}  & 
  FT & 41.66 & 26.04 & 17.86 & 12.91 & 70.2 & 34.03 \\
& PT1+FT & 40.71 & 27.61 & 20.48 & 15.78 & 94.6 & 37.09 \\
& PT1+PT2+FT & 40.5 & 28.6 & 21.67 & 16.93 & 92.1 & 38.51 \\
& PT1+PT2+ITA+FT & 39.57 & 27.78 & 21.05 & 16.48 & 101.2 & 37.71 \\
& PT1+PT2+ITA+SEL+FT & 40.39 & 28.46 & 21.57 & 16.86 & 103.9 & 38.45 \\

\midrule
\multirow{5}{*}{NusceneMQA}  &
  FT & 22.39 & 15.53 & 10.51 & 6.73 & 34.4 & 19.07 \\
& PT1+FT & 52.26 & 42.92 & 34.18 & 26.07 & 21.8 & 51.19 \\
& PT1+PT2+FT & 55.24 & 44.99 & 35.42 & 26.75 & 23.05 & 52.02  \\
& PT1+PT2+ITA+FT & 56.44 & 45.83 & 36.18 & 27.53 & 23.45 & 53.29\\
& PT1+PT2+ITA+SEL+FT & 67.4 & 60.42 & 52.98 & 46.81 & 32.01 & 67.47 \\

\midrule		
\multirow{5}{*}{NuInstruct}  & 
  FT & 50.43 & 44.63 & 43.1 & 39.19 & 333.34 & 59.8 \\
& PT1+FT & 51.49 & 49.73 & 48.08 & 46.7 & 384.74 & 66.63 \\
& PT1+PT2+FT & 56.86 & 55.2 & 53.69 & 52.44 & 406.9 & 71.73 \\
& PT1+PT2+ITA+FT & 66.87 & 65.78 & 64.62 & 63.65 & 704.15 & 82.69 \\
& PT1+PT2+ITA+SEL+FT & 74.7 & 73.11 & 71.41 & 69.85 & 672.5 & 83.63 \\

\bottomrule
\end{tabular}
\label{tab:detail_result}
}
\end{table*}

\section{Visualization}
\begin{figure*}
    \centering
    \includegraphics[width=\linewidth]{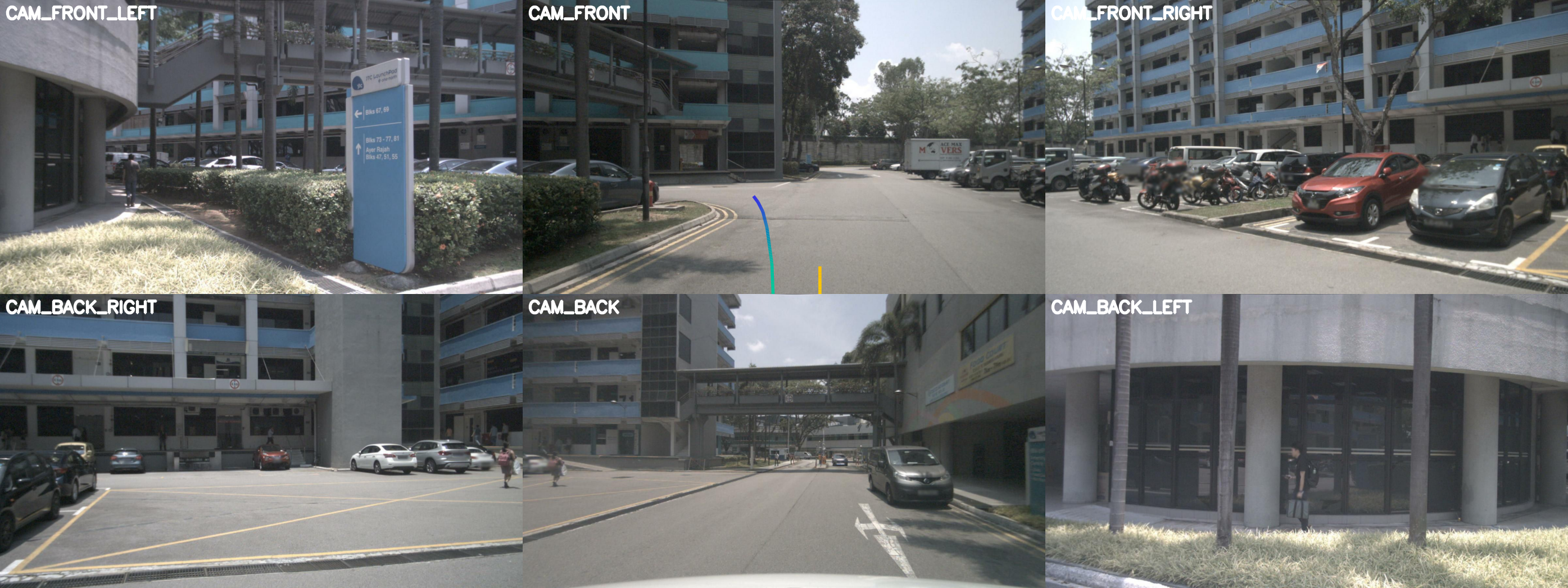}
    \includegraphics[width=\linewidth]{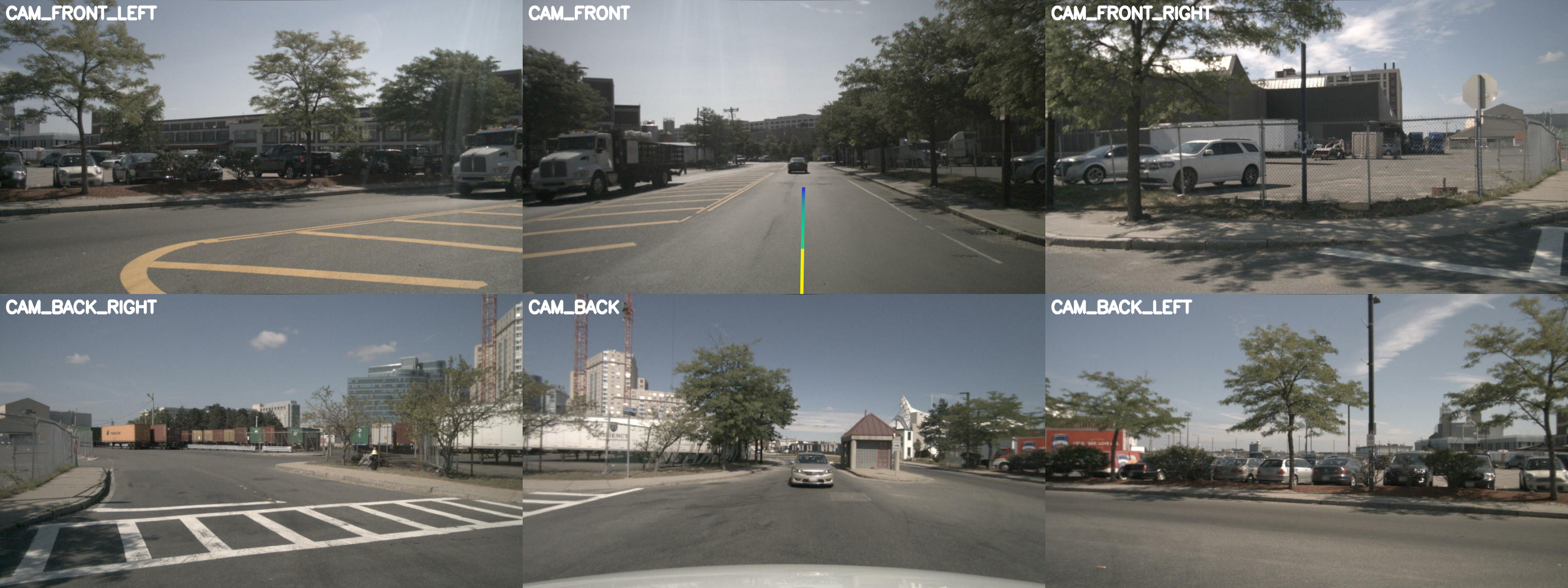}
    \includegraphics[width=\linewidth]{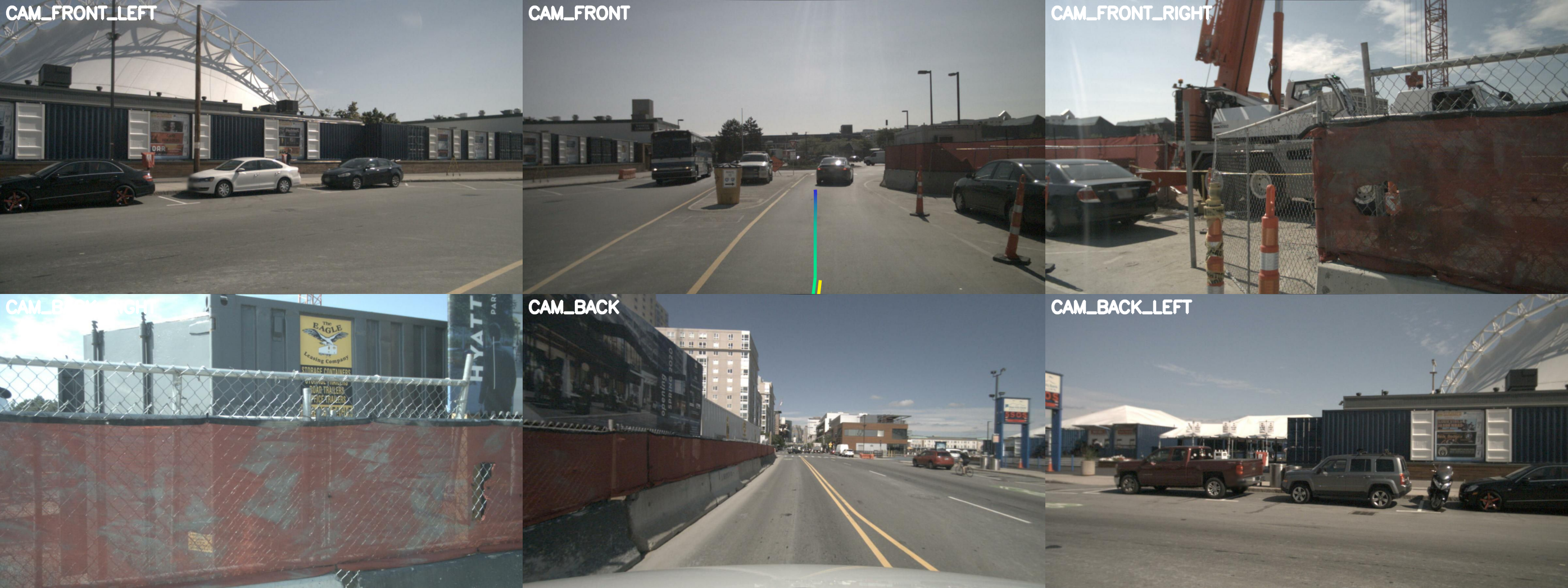}
    \caption{Bad Case Visualization}
    \label{fig:bad_case}
\end{figure*}

\paragraph{Bad Case Analyse.} 
As shown in Figure~\ref{fig:bad_case}, a detailed visualization and analysis of several challenging cases highlight that the scenarios associated with higher L2 errors are primarily encountered in more open environments. In these situations, our model demonstrates a tendency to accelerate, whereas the ground truth planning typically opts for deceleration. This divergence in behavior suggests a potential area for improvement. However, we posit that the acceleration strategy employed by our model is not only deliberate but also more reasonable under the given circumstances. This is because, in open environments, maintaining or increasing speed can often be a safer and more efficient approach, aligning better with the overarching goals of our planning algorithm.

\section{Experiments}
The comprehensive metrics presented in Table~\ref{tab:detail_result} clearly demonstrate the effectiveness of our proposed training strategy, interactor module, and selection operation across all three datasets. These results highlight the consistent performance improvements achieved through our approach. However, it is important to note that the single-view pre-training stage does not contribute significantly to enhancing language-related scores. We attribute this to the design of the single-view pre-training stage, which is primarily focused on refining object perception capabilities, thus providing a robust foundation for tackling the multi-view grounding task rather than directly influencing language understanding.

\paragraph{Token Redundancy.}
Intuitively, certain visual patch tokens, such as those representing sky regions, may be irrelevant or weakly related to the instruction. To support this, we included an additional experiment showing the potential redundancy of tokens in specific contexts. We randomly masked visual tokens corresponding to sky regions, testing various masking rates. As shown in table~\ref{tab:mask}, "blind" indicates replacing the visual input with random values in Exp.1. Comparing Exp.1 and Exp.2 demonstrates the role of visual features. The minimal performance change between Exp.3 to Exp.4 and Exp.2 indicates that some redundant tokens in the visual input are indeed irrelevant to language instructions. The substantial performance drop of Exp.5 shows that critical information was masked.

\begin{table*}[htb]
    \centering
    \caption{Comparison of model performance under different visual mask rates.}
    \begin{tabular}{cccccc}
    \toprule
         Exp. & Mask Rate ($\%$) & MEA $\downarrow$ & ACC $\uparrow$ & mAP $\uparrow$ & BLUE $\uparrow$ \\
    \midrule
         1 & blind  & 15.6&	25.14&	0&	    25.3 \\
         2 & 0      & 4.33&	52.71&	16.66&	69.85 \\
         3 & 10     & 4.43&	51.08&	16.49&	69.91 \\
         4 & 30     & 3.8&	52.88&	15.4&	68.73 \\
         5 & 50     & 4.63&	46.24&	11.62&	63.65 \\
    \bottomrule
    \end{tabular}
    \label{tab:mask}
\end{table*}

\paragraph{BEV Encoder.} 
We argue that using future frames is unreasonable, and to more accurately reflect the performance of our method, we did not fully adhere to the training setup of SparseBEV (i.e., using validation set data for training). 
Consequently, we retrained SparseBEV. 
As shown in table~\ref{tab:sparsebev}, our retrained SparseBEV is compared with the original model.

\begin{table*}[htb]
    \centering
    \caption{The results of SparseBEV. $\dagger$ means without validation set data. $\ddagger$ means withou future frames.}
    \begin{tabular}{cccc}
    \toprule
         MODEL & BACKBONE & NDS & mAP \\
    \midrule
         SparseBEV & ResNet50 & 54.5 & 43.2 \\
         SparseBEV & ResNet50 & 55.8 & 44.8 \\
         SparseBEV & ResNet101 & 59.2 & 50.1 \\
    \midrule
         SparseBEV & ViT & 85.3 & 86.71 \\
    \midrule
         SparseBEV $\dagger$ & ViT & 67.75 & 61.28 \\
         SparseBEV $\ddagger$ & ViT & 64.66 & 56.82 \\
    \bottomrule
    \end{tabular}
    \label{tab:sparsebev}
\end{table*}

\section{Discussion}
Recent advancements in Multi-modal Large Language Models (MLLMs) have demonstrated significant capabilities in multi-image scene understanding~\cite{awadalla2023openflamingo,laurenccon2024matters,lin2024vila,jiang2024mantis,sun2024generative}.
Approaches for handling multiple image inputs can be broadly categorized into two types: 
(1) concatenating multiple images into a single image and feeding it into the large language model, and 
(2) extracting features from each image individually and concatenating these features before inputting them into the large language model. 
The former approach significantly reduces the resolution of the input images, leading to a loss of image details. The latter approach substantially increases the input sequence length, which may exceed the maximum input length of the large language model when the number of images exceeds six.
Our model adopts a novel approach where relevant features are extracted based on user instructions, and potential details lost are supplemented from the original features. This method leverages the strengths of the aforementioned approaches, maintaining detailed features while keeping the input sequence length within acceptable limits.

Although our method demonstrates impressive performance in multi-image scene understanding and achieves comparable results in open-loop driving scenarios, it has not yet been tested on closed-loop datasets such as Carla~\cite{Dosovitskiy17} or NuPlan~\cite{nuplan}. Additionally, the method has not been validated for 3D Grounding tasks due to the reliance on precise camera parameters for 2D-to-3D conversion, which are challenging to tokenize. We will address these issues in future work.

\begin{table*}
    \centering
    \caption{The first step in generating data involves using examples of prompts and responses. Given the objects in the scene, use GPT-4o to extract risks associated with each object.}
    \begin{tabular}{cp{15cm}}
        \textbf{Prompt} & The image is from the front view camera of ego vehicle, and please provide a risk assessment of the given object to ego vehicle. The driving risk categories include: 1. View obstruction. 2. Collision possibility. 3. Traffic rule violations. 4. Potential risk. You are now a driver, and from the perspective of driving safety, you need to conduct a driving risk analysis.Please consider the state of the target when analyzing, e.g. Whether the vehicle is stationary, whether pedestrians are crossing the road, whether it is in the same lane as ego vehicle, etc. The current scene contains the following objects: [the car located 26 meters ahead to the right; the car located 26 meters ahead to the right; the car located 26 meters ahead to the left; the car located 16 meters ahead to the left; the car located 26 meters ahead to the left; the car located 26 meters ahead to the right; the car located 23 meters ahead]. Choose the object you believe poses a risk and provide your reasons. If all risks of object are None, ignore this object! If some risk is None, do not output all context relate to this risk! Answer in the following format without providing additional information:
        \begin{verbatim}
{
    "[obj]": {
        "View obstruction": {
            "Status": "[High/Medium/Low/None]", 
            "Reason": "[Reason]"
        },
        "Collision possibility": {
            "Status": "[High/Medium/Low/None]", 
            "Reason": "[Reason]"
        }, 
        ...
        }, 
    "[obj]": {
        ...
    }, 
    ...
} 
        \end{verbatim}
        \\
        \textbf{Response} & 
        \begin{verbatim}
{
    "the car located 16 meters ahead to the left": {
        "Collision possibility": {
            "Status": "High",
            "Reason": "It is in the turning path of the ego vehicle and 
            poses a risk of collision."
        },
        "Potential risk": {
            "Status": "High",
            "Reason": "Being in close proximity and suggesting movement 
            or turning, it presents a potential risk."
        }
    }
}
        \end{verbatim}
    \end{tabular}
    \label{tab:extract_risk}
\end{table*}

\newpage

\begin{table*}
    \centering
    \caption{The second step in generating data involves using examples of prompts and responses. we use GPT-4o-mini to organize the object-level risks generated in the first step into a rich QA format.}
    \begin{tabular}{cp{15cm}}
    \textbf{Prompt} & This is a description of object-level traffic risks: 1. car causes low view obstruction risk due to the car is parked on the side of the road and does not obstruct the view of the ego vehicle. 2. car causes low collision possibility risk due to the car is stationary and parked on the side of the road, not in the path of the ego vehicle. Please generate multiple Q\&A pairs about traffic risks based on this information and output them in JSON format as follows:
        \begin{verbatim}
 [
     {
        "question": [question1], 
        "answer": [answer1]},
     {
        "question": [question2], 
        "answer": [answer2]
     },
     ...
 ]
        \end{verbatim}\\
        \textbf{Response} & 
        \begin{verbatim}
[
    {
        "question": "What type of traffic risk is presented by a 
        parked car on the side of the road?", 
        "answer": "The parked car presents a low view obstruction 
        risk to the ego vehicle as it does not obstruct their view."
    },    
    {
        "question": "Does a stationary parked car cause a risk of 
        collision for the ego vehicle?", 
        "answer": "No, the stationary parked car causes a low colli-
        sion possibility risk because it is not in the path of the 
        ego vehicle."
    },    
    {
        "question": "How does a parked car affect the view of the 
        ego vehicle?", 
        "answer": "A parked car causes low view obstruction risk, 
        meaning it does not significantly obstruct the view of theego
        vehicle."
    },
    {
        "question": "Why is the collision risk low when a car is parked 
        on the side of the road?", 
        "answer": "The collision risk is low because the car is station-
        ary and not in the path of the ego vehicle."
    },
    {
        "question": "What factors contribute to the low collision risk 
        of a parked vehicle?", 
        "answer": "The main factors are that the car is stationary and 
        parked on the side of the road, thus not interfering with the 
        ego vehicle's path."
    },
    {
        "question": "Can a parked car pose any significant risks to tra-
        ffic safety?",
        "answer": "Generally, a parked car poses low risks such as low 
        view obstruction and low collision possibility for actively mov-
        ing vehicles."
    }
]
        \end{verbatim} \\
    \end{tabular}
    \label{tab:extract_qa}
\end{table*}

\end{document}